\documentclass{article}
\usepackage{spconf,amsmath,graphicx}
\usepackage[utf8]{inputenc} 
\usepackage[T1]{fontenc}    
\usepackage{hyperref}       
\usepackage{url}            
\usepackage{booktabs}       
\usepackage{amsfonts}       
\usepackage{nicefrac}       
\usepackage{microtype}      
\usepackage{xcolor}         
\usepackage[numbers]{natbib}
\usepackage{amsmath}
\usepackage{svg} 
\usepackage{graphicx}
\usepackage{bm}
\usepackage{color}
\usepackage{hyperref}
\usepackage{tabularray}
\usepackage{algorithm}
\usepackage{algorithmicx} 
\usepackage{algpseudocode}
\usepackage{pifont}
\usepackage{bbding} 
\usepackage[normalem]{ulem}

\hypersetup{
    colorlinks=true,
    urlcolor=magenta,
}

\title{SAM-Deblur: Let Segment Anything Boost Image Deblurring}
\name{Siwei Li$^{\star \mathsection}$,
Mingxuan Liu$^{\dagger}$,
Yating Zhang$^{\star}$,
Shu Chen$^{\star}$,
Haoxiang Li$^{\dagger}$, 
Zifei Dou$^{\star}$,
Hong Chen$^{\ddagger}$
\thanks{Siwei Li and Mingxuan Liu are co-first authors of the article.}
\thanks{This work is supported by the National Science and Technology Major Project from Minister of Science and Technology, China (Grant No. 2018AAA0103100), and National Natural Science Foundation of China (No. 92164110, 62334014 and U19B2041), partly supported by Beijing Engineering Research Center (No. BG0149)(Corresponding authors: Hong Chen and Zifei Dou.)}}

 \address{$^{\star}$ Beijing Xiaomi Mobile Software Co., Ltd. \\
    $^{\mathsection}$ Department of Electronic Engineering, Tsinghua University\\
      $^{\dagger}$ Department of Biomedical Engineering, Tsinghua University \\
      $^\ddagger$ School of Integrated Circuits, Tsinghua University} 

\begin{document}
\topmargin=0mm
%
\maketitle
\begin{abstract}
Image deblurring is a critical task in the field of image restoration, aiming to eliminate blurring artifacts. However, the challenge of addressing non-uniform blurring leads to an ill-posed problem, which limits the generalization performance of existing deblurring models. To solve the problem, we propose a framework SAM-Deblur, integrating prior knowledge from the Segment Anything Model (SAM) into the deblurring task for the first time. In particular, SAM-Deblur is divided into three stages. First, we preprocess the blurred images, obtain segment masks via SAM, and propose a mask dropout method for training to enhance model robustness.  Then, to fully leverage the structural priors generated by SAM, we propose a Mask Average Pooling (MAP) unit specifically designed to average SAM-generated segmented areas, serving as a plug-and-play component which can be seamlessly integrated into existing deblurring networks. Finally, we feed the fused features generated by the MAP Unit into the deblurring model to obtain a sharp image. Experimental results on the RealBlurJ, ReloBlur, and REDS datasets reveal that incorporating our methods improves GoPro-trained NAFNet's PSNR by 0.05, 0.96, and 7.03, respectively. Project page is available at GitHub \href{https://hplqaq.github.io/projects/sam-deblur}{HPLQAQ/SAM-Deblur}.
\end{abstract}
\section{Introduction}

\begin{figure}[htb]
\centering
\includegraphics[width=0.45\textwidth]{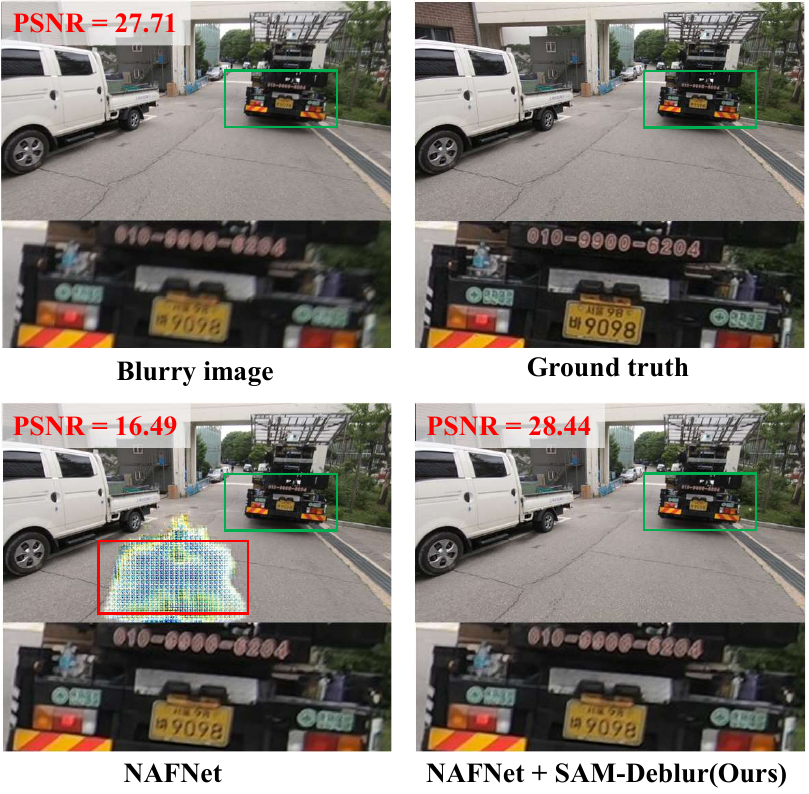}
\caption{The "mode collapse" in NAFNet: NAFNet trained on the GoPro dataset \cite{d6} may output anomalous pixel regions during testing on the REDS dataset \cite{reds} (indicated by the red boxes). This problem can be effectively solved by the proposed SAM-Deblur.}
\label{fig1}
\end{figure}

When capturing images, factors like camera shake, defocusing, and high-speed motion can cause image blurring, impacting various practical applications \cite{d1}. Developing image deblurring models to transform blurred images into sharp ones can greatly improve downstream tasks such as object detection, autonomous driving, and visual navigation\cite{d2}. Therefore, image deblurring is a vital task in computer vision.

Deep learning has excelled in image deblurring tasks \cite{d3,d4,db1,db2,db3}, with models like MIRNet-v2 \cite{d3} and Uformer \cite{d4} setting new benchmarks. MIRNet-v2 integrates multi-scale contextual information, while Uformer, a Transformer-based model, optimizes computational efficiency for high-resolution images. These models achieve impressive PSNR scores on benchmarks like SIDD \cite{d5} and GoPro \cite{d6}. However, they are computationally expensive. To address this issue, Chen et al. introduced NAFNet \cite{d7}, a streamlined and nonlinear activation free network that maintains state-of-the-art (SOTA) performance while reducing computational overhead.

\begin{figure}[htb]
\centering
\includegraphics[width=0.48\textwidth]{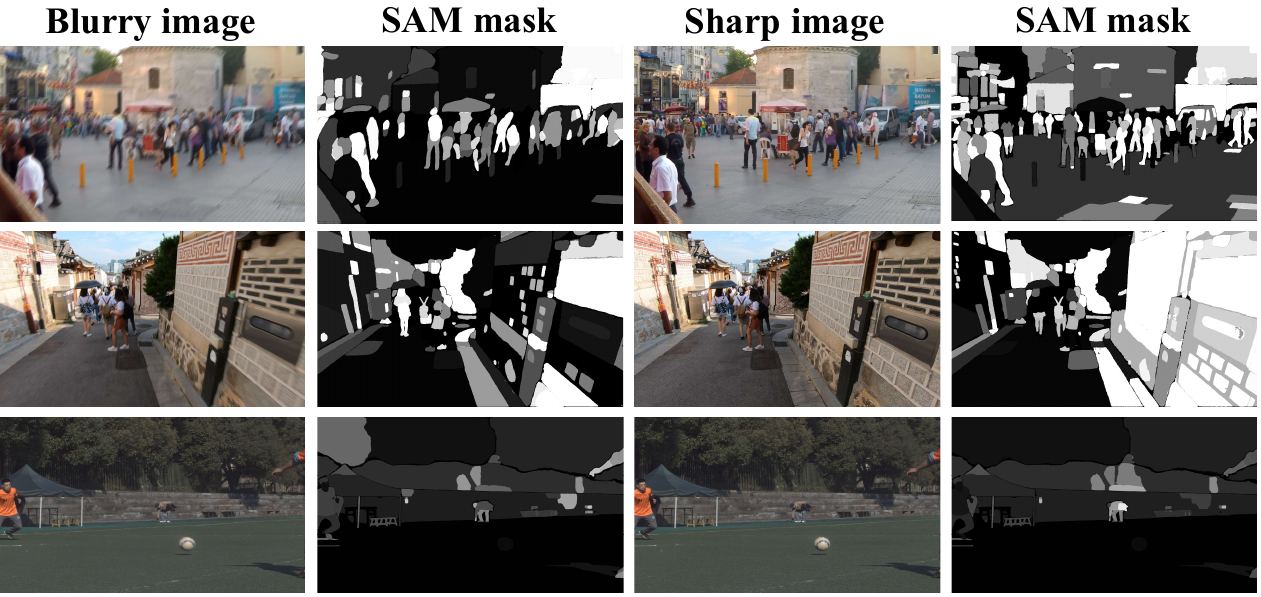}
\caption{Illustration of SAM’s robustness on blurred images.}
\label{fig2}
\end{figure}

However, image deblurring is an ill-posed problem with a large solution space for non-uniform blur \cite{d1}, limiting model generalization. For example, the SOTA deblurring model NAFNet with 32 channels, trained on the GoPro dataset, can yield a high PSNR of 32.85 on in-distribution (ID) test sets. Nevertheless, its performance deteriorates on out-of-distribution (OOD) datasets, even experiencing "mode collapse" (shown in Fig. \ref{fig1}). This severely limits the practical potential of deblurring models.

In this paper, to regularize the solution space of latent sharp images in deblurring tasks and thereby improve the generalization performance of existing models, we pioneer the use of semantic-aware priors from the pretrained Segment Anything Model (SAM) \cite{d8} to regularize the solution space in deblurring tasks. The motivation for using SAM as a semantic prior is based on its robustness to blurred images, which is due to SAM's pretraining on the expansive SA-1B dataset \cite{d8}, encompassing a staggering 10 billion masks and 11 million images. As shown in Fig. \ref{fig2}, even under suboptimal image quality, SAM succeeds in accurately segmenting most of objects. 

Some studies have applied SAM priors to image restoration tasks like dehazing \cite{hazing} by using grayscale coding and channel expansion, highlighting their utility in low-level visual tasks. However, due to the complex causes and larger solution space for non-uniform blur in deblurring, these methods are not directly transferable. To better adapt SAM for the task of deblurring, we propose a framework, SAM-Deblur, which for the first time incorporates SAM-generated priors to enhance image deblurring. Specifically, SAM-Deblur consists of three main stages. First, we preprocess blurred images and use SAM to obtain segment masks. For model training, we propose a mask dropout method during preprocessing to improve model robustness. Then, we propose a plug-and-play Mask Average Pooling (MAP) Unit designed for SAM, which averages segmented regions and can be readily integrated into existing deblurring networks. Finally, we use the fused features created by the MAP Unit as input to the deblurring model to achieve a sharp image. Extensively experiment on RealBlurJ \cite{realblur}, REDS \cite{reds}, and ReLoBlur \cite{relo} show that SAM priors improve deblurring performance and mitigate "mode collapse".

\begin{algorithm}
\label{alg1}
\caption{Mask Average Pooling (MAP) Unit}
\begin{algorithmic}[1]
\Require Input image tensor \(I_{bm} \in \mathbb{R}^{H \times W \times C_{in}}\), Mask tensor \(I_{mask} \in \{ 0,1 \} ^{H \times W \times N_{m}}\), Initialize \( I_{\text{MAP}} \) to zeros

\State \( I_{\text{enc}} \in \mathbb{R}^{H \times W \times S} \gets \text{ Encoder }(I_{bm}) \)

\State \textbf{If} \( \text{training} \) \textbf{then} \( \text{Dropout}(I_{mask}) \) \Comment{Mask Dropout}

\State \( I_{mask} \gets \text{Concat}(I_{mask}, mask_{uncovered}) \)

\For{ \( mask_i \in I_{mask} \) } \Comment{Mask Average Pooling}
    \State \( I_{\text{enc\_i}} \gets I_{\text{enc}} \times mask_i \)
    \State \( I_{\text{MAP\_i}} \gets \text{Average \ Pooling}(I_{\text{enc\_i}}) \)
    \State \( I_{\text{MAP}} \gets I_{\text{MAP}} + I_{\text{MAP\_i}} \)
\EndFor

\State \(I_{input} \gets \text{Concat} (I_{\text{MAP}}, I_{bm})\)

\Ensure Output tensor \( I_{input} \in \mathbb{R}^{H \times W \times (S+C_{in})} \)
\end{algorithmic}
\end{algorithm}
\section{Preliminary}
\subsection{Network Definition}
\begin{figure*}[htb]
\centering
\includegraphics[width=1\textwidth]{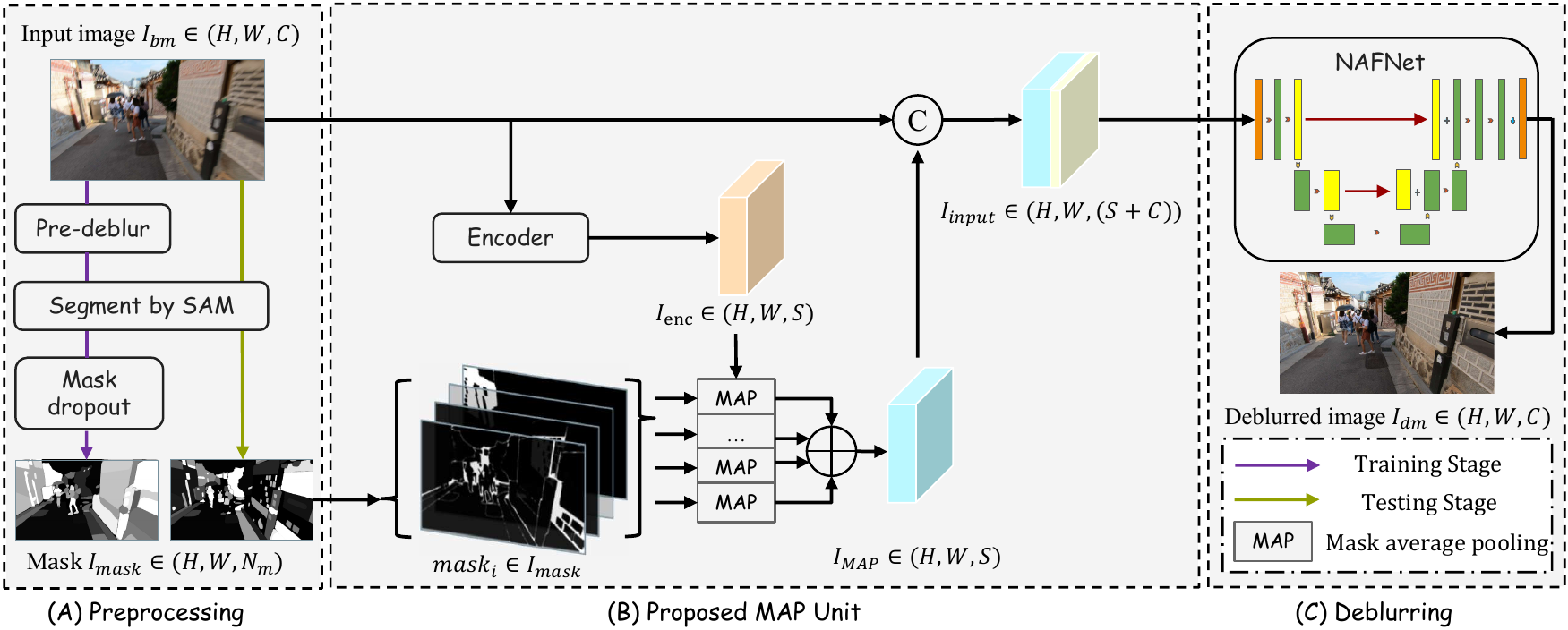}
\caption{Illustration of SAM-Deblur framework. (A) Preprocessing: In training, we first deblur $I_{bm}$ using GoPro-pretrained NAFNet, then segment it with SAM and finally apply our mask dropout to get $I_{mask}$. In testing, $I_{dm}$ is directly fed into SAM for segmentation. (B) Proposed MAP Unit: $I_{dm}$ and $I_{mask}$ are then processed by a MAP Unit to obtain the SAM prior, which is concatenated with $I_{dm}$ to get $I_{input}$ as input of NAFNet. (C) Deblurring: NAFNet implements image deblurring based on the SAM prior contained in $I_{input}$. }
\label{fig3}
\end{figure*}

For a blurred image $I_{bm} \in \mathbb{R}^{H \times W \times C_{in}}$, the deblurring model NAFNet Net($\cdot$) can generate a sharp image $I_{dm} \in \mathbb{R}^{H \times W \times C_{out}}$, where the $H$, $W$, $C_{in}$, and $C_{out}$ are the image height, width, input and output channel. As illustrate in Fig. \ref{fig3}(C), NAFNet employs a classical single-stage U-shaped architecture with skip connections to mitigate inter-block complexity \cite{unet}. Additionally, NAFNet utilizes nonlinear activation-free network blocks, in which Channel Attention (CA) and GELU are replaced with Simplified Channel Attention (SCA) and SimpleGate, respectively.

\subsection{SAM as Structural Prior for Deblurring}
In image restoration, deblurring is often formulated as solving the inverse problem defined by the equation $Y = H \otimes X + \epsilon $. Here, $Y$ denotes the observed blurred image, $X$ is the latent sharp image we aim to recover, $H$ represents the blur kernel affecting the image, $\epsilon$ signifies measurement noise, and $\otimes$ stands for the convolution operation. To address this problem, we employ a deblurring model, $\text{F}$, designed to approximate $X = \text{F}(Y) $. A particular challenge in this task is that the blur kernel $H$ can vary in different local regions within the image, necessitating that the model $\text{F}$ has the ability to adaptively aggregate local features for appropriate image recovery.

SAM is a highly-parameterized model, extensively trained on a corpus of 1 billion masks extracted from 11 million images. This extensive training enables SAM to achieve robust performance in segmenting almost all objects in images, regardless of the image quality. Given an input image \(I \in \mathbb{R}^{H \times W \times C_{in}}\), SAM is capable of generating a segmentation mask \(I_{mask} \in \{0,1\}^{H \times W \times{N_{m}}}\), where \(N_{m}\) is the number of generated masks. Inspired by these capabilities, we leverage the structural priors offered by SAM to tackle the complex issue of localized blurring in our deblurring tasks.

\section{Methodology}
\subsection{SAM-Deblur Pipeline}
The SAM-Deblur pipeline is visualized in Fig. \ref{fig3}. In training, given an input image \(I_{bm} \in \mathbb{R}^{C \times H \times W}\), we use GoPro-pretrained NAFNet to pre-deblur \(I_{bm}\) into a sharp image. Then SAM generates masks $I_{mask}$ from the sharp image, represented as a Boolean tensor \(I_{mask} \in \{0,1\}^{H \times W \times N_{m}}\). Each mask of size \(H \times W\) identifies a specific region in the image with values set to 1. After that, the input image \(I_{bm}\) and the generated masks \(I_{mask}\) are fed into MAP Unit to integrate the blur priors of the image and the structural priors obtained from the semantic masks generated by SAM. Finally, the output from the MAP Unit $I_{input} \in \mathbb{R}^{H \times W \times (S+C_{in})}$ will serve as the input for NAFNet. During testing, we directly use SAM to segment the original image and keep all other processes consistent with the training phase. 

\subsection{Mask Average Pooling Unit}
 The core component of our proposed method is the MAP Unit, which is described in detail in Algorithm \ref{alg1}. MAP facilitates a rich interaction between image information and the segmentation masks generated by SAM, producing localized blur priors that are subsequently channel-concatenated with the original image and fed into the NAFNet.

As depicted in Fig. \ref{fig3}(B), the MAP Unit takes two inputs: the blurred image $I_{bm}$ and the mask $I_{mask}$. The unit first extracts features $I_{enc} \in \mathbb{R}^{H \times W \times S}$ from the image through an encoder, which is designed as a combination of a \(3 \times 3\) convolution followed by a \(1 \times 1\) convolution and is co-trained with the deblurring network.

After image encoding, the MAP operation is performed. For each mask $mask_i \in \mathbb{R} ^{H \times W \times 1}$, we multiply it with the image encoding $I_{enc}$ and calculate the average of $I_{enc}$ within the region defined by the mask. The average value is then reassigned to this region. As a result, the MAP Unit produces a fusion of image features and SAM priors, represented as $I_{{\rm{MAP}}} \in \mathbb{R}^{H \times W \times S}$, which aggregates blended prior information at each pixel from its corresponding region.

Quality of the generated masks can vary, influenced by factors such as the size of the SAM model and the level of blur present in the image. To prevent the model from over-relying on specific regional information, we introduce a mask dropout strategy shown in Fig. \ref{fig1}(A). During training, some of the masks are randomly discarded, and the regions they cover are eventually aggregated into the background.

\begin{table*}
\label{tb1}
\caption{Quantitative evaluation results. Best results are highlighted in bold. (w/o SAM: Not using SAM priors, CAT \cite{hazing}: concatenation method proposed in \cite{hazing}, MAP: Using SAM-Deblur framework w/o mask dropout, Ours: Using SAM-Deblur framework.)}
\centering
\begin{tblr}{
  cell{1}{1} = {r=2}{},
  cell{1}{2} = {c=2}{},
  cell{1}{4} = {c=3}{},
  cell{1}{7} = {c=3}{},
  cell{1}{10} = {c=3}{},
  hline{1,3,7} = {-}{},
  hline{2} = {2-12}{},
}
Methods           & GoPro          &                 & RealBlurJ      &                 &                 & REDS           &                 &                 & ReLoBlur       &                 &                  \\
                     & PSNR↑          & SSIM↑           & PSNR↑          & SSIM↑           & MCR↓            & PSNR↑          & SSIM↑           & MCR↓            & PSNR↑          & SSIM↑           & MCR↓             \\
w/o SAM                 & 32.85          & 0.960         & 26.57          & 0.866          & 0.20\%          & 25.97          & 0.844          & 3.80\%          & 25.26          & 0.687          & 54.68\%          \\ CAT \cite{hazing}               & \textbf{32.88} & \textbf{0.961} & 26.55          & 0.863          & 0.31\%          & 26.65          & 0.865          & 2.57\%          & 29.77          & 0.882          & 58.73\%          \\
MAP                 & 32.82          & 0.960          & 26.57          & 0.866         & 0.31\%          & 26.81          & 0.865         & 0.40\%          & 30.86          & 0.897          & 55.44\%          \\
\textbf{Ours} & 32.83          & 0.960          & \textbf{26.62} & \textbf{0.867} & \textbf{0.00\%} & \textbf{26.93} & \textbf{0.868} & \textbf{0.20\%} & \textbf{32.29} & \textbf{0.903} & \textbf{13.92\%} 
\end{tblr}
\end{table*}

\begin{figure}[htb]
\centering
\includegraphics[width=0.48\textwidth]{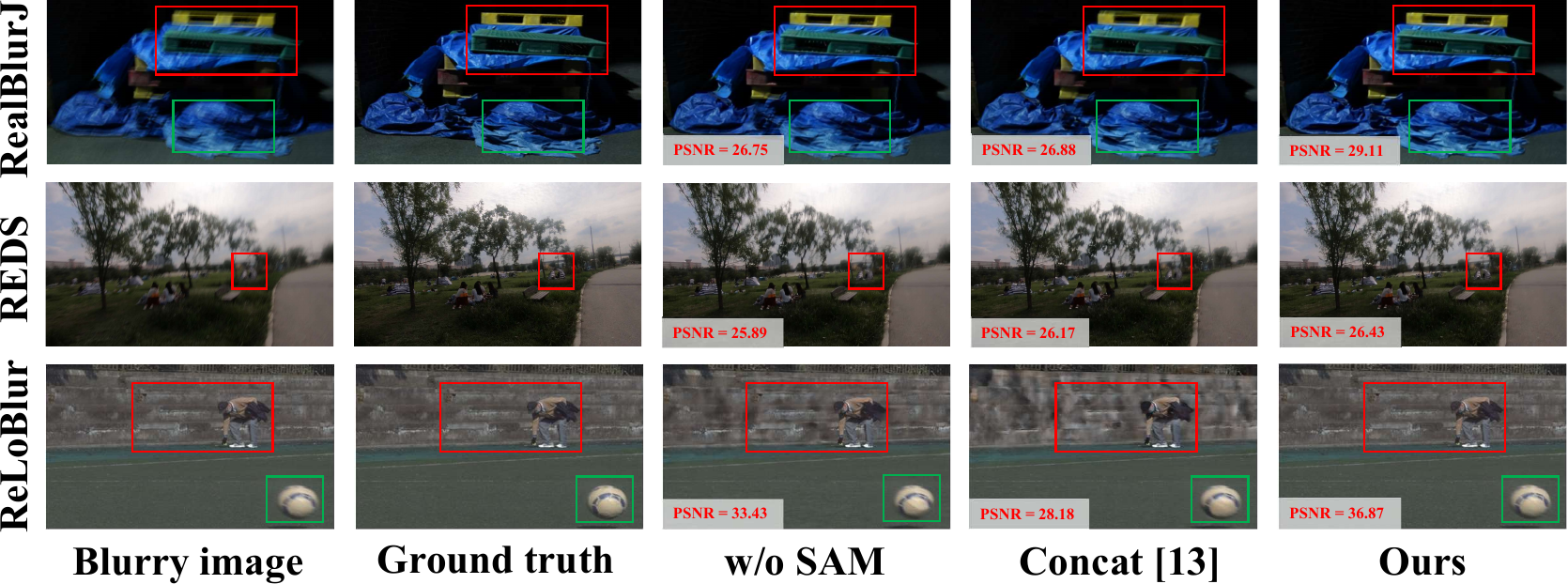}
\caption{ Illustration of our deblurred images on RealBlurJ, REDS, and ReLoBlur. Please zoom in for a better review. }
\label{fig4}
\end{figure}

\section{Experiments}

\subsection{Datasets}
For the training phase, we employ the GoPro dataset \cite{d6} which consists of 2,103 training images and 1,111 test images. To rigorously assess the model's generalization capabilities, we further test the NAFNet trained with various methods on three widely-recognized real-world datasets with data distributions differing from GoPro. Specifically, we utilize RealBlurJ \cite{realblur} with 980 images, REDS \cite{reds} comprising 3,000 images, and ReLoBlur \cite{relo}, which contains 395 images.

\subsection{Experimental Setup}

\textbf{Compared methods:} We utilize NAFNet with 32 channels as the image deblurring model and compare four different training methods: one without using SAM priors, another employing the concatenation method for introducing SAM priors as proposed in \cite{hazing}, a third method using only the MAP Unit, and finally, our proposed SAM-Deblur framework that combines both the MAP Unit and mask dropout. In addition, TLC \cite{tlc} is only used when testing on GoPro.

\textbf{Metrics:} For both the GoPro test set and the three OOD real-world datasets, we employ two commonly used metrics: Peak Signal-to-Noise Ratio (PSNR) and Structural SIMilarity (SSIM) \cite{psnr}. Higher values for these metrics indicate better deblurring performance. Additionally, we have observed that NAFNet trained on the GoPro exhibits "mode collapse" when tested on OOD datasets (Fig. \ref{fig1}). To quantify this, we introduce the Mode Collapse Rate (MCR), calculated using a threshold-based method. Specifically, when \( \text{PSNR}(I_{bm}, I_{gt}) - \text{PSNR}(I_{dm}, I_{gt}) > 3 \) ($I_{gt}$ is the ground truth), we consider the model to have undergone "mode collapse". A lower MCR suggests stronger generalization capabilities of the model.

\textbf{Training Details:} All models are trained using the Adam optimizer with \( \beta_1 = 0.9 \), \( \beta_2 = 0.9 \), and a weight decay of 0.001. Initial learning rate is 0.001 and gradually reduced to \(1 \times 10^{-6}\) using cosine annealing. We adopt a training patch size of \(256 \times 256\), a batch size of 32, and train for a total of 200k iterations. Data augmentation techniques such as shuffling, rotation, and flipping are applied during training.

\subsection{Results and Discussion}
The quantitative experimental results are presented in Table 1. On the ID test set GoPro, the differences of results among the four training methods are relatively minor. Specifically, the PSNR varies by less than 0.05, and the SSIM varies by less than 0.001. On the OOD test sets RealBlurJ, REDS, and ReLoBlur, our proposed method achieves the best deblurring performance. Compared to the training method without SAM priors, the PSNR improved by 0.05, 0.96, and 7.03, while the SSIM increased by 0.001, 0.024, and 0.216, respectively. The MCR also decreased by 0.20\%, 2.6\%, and 40.76\%, respectively. Moreover, we observe that the method proposed in \cite{hazing} does not perform well in the utilization of SAM priors, resulting in inferior test results on all three OOD datasets compared to our proposed method. Qualitative results are shown in Fig. \ref{fig4}, indicating that the SAM-Deblur framework achieves better deblurring performance on OOD real-world datasets.

\section{Conclusion}
We propose a framework, SAM-Deblur, which can boost image deblurring with semantic-aware priors from the Segment Anything Model (SAM). The core of SAM-Deblur is the proposed Mask Average Pooling (MAP) unit, which enables a robust interaction between image data and segmentation masks generated by SAM. Alongside MAP, we introduce a mask dropout method to enhance robustness of the MAP unit in handling the variability of masks generated by SAM. Experiments on RealBlurJ, REDS, and ReLoBlur, show that our proposed method can improve generalizability of the deblurring model NAFNet and mitigate "mode collapse". Our method can open new avenues for the practical application of deblurring algorithms in real-world scenarios.

\bibliographystyle{IEEEbib}
\bibliography{Template}

\begin{thebibliography}{10}

\bibitem{d6}
Seungjun Nah, Tae Hyun~Kim, and Kyoung Mu~Lee,
\newblock ``Deep multi-scale convolutional neural network for dynamic scene
  deblurring,''
\newblock in {\em Proceedings of the IEEE conference on computer vision and
  pattern recognition}, 2017, pp. 3883--3891.

\bibitem{reds}
Seungjun Nah, Sungyong Baik, Seokil Hong, Gyeongsik Moon, Sanghyun Son, Radu
  Timofte, and Kyoung~Mu Lee,
\newblock ``Ntire 2019 challenge on video deblurring and super-resolution:
  Dataset and study,''
\newblock in {\em CVPR Workshops}, June 2019.

\bibitem{d1}
Jiang Hai, Ren Yang, Yaqi Yu, and Songchen Han,
\newblock ``Combining spatial and frequency information for image deblurring,''
\newblock {\em IEEE Signal Processing Letters}, vol. 29, pp. 1679--1683, 2022.

\bibitem{d2}
Kaihao Zhang, Wenqi Ren, Wenhan Luo, Wei-Sheng Lai, Bj{\"o}rn Stenger,
  Ming-Hsuan Yang, and Hongdong Li,
\newblock ``Deep image deblurring: A survey,''
\newblock {\em International Journal of Computer Vision}, vol. 130, no. 9, pp.
  2103--2130, 2022.

\bibitem{d3}
Syed~Waqas Zamir, Aditya Arora, Salman Khan, Munawar Hayat, Fahad~Shahbaz Khan,
  Ming-Hsuan Yang, and Ling Shao,
\newblock ``Learning enriched features for fast image restoration and
  enhancement,''
\newblock {\em IEEE transactions on pattern analysis and machine intelligence},
  vol. 45, no. 2, pp. 1934--1948, 2022.

\bibitem{d4}
Zhendong Wang, Xiaodong Cun, Jianmin Bao, Wengang Zhou, Jianzhuang Liu, and
  Houqiang Li,
\newblock ``Uformer: A general u-shaped transformer for image restoration,''
\newblock in {\em Proceedings of the IEEE/CVF conference on computer vision and
  pattern recognition}, 2022, pp. 17683--17693.

\bibitem{db1}
Kiyeon Kim, Seungyong Lee, and Sunghyun Cho,
\newblock ``Mssnet: Multi-scale-stage network for single image deblurring,''
\newblock in {\em European Conference on Computer Vision}. Springer, 2022, pp.
  524--539.

\bibitem{db2}
Xintian Mao, Yiming Liu, Fengze Liu, Qingli Li, Wei Shen, and Yan Wang,
\newblock ``Intriguing findings of frequency selection for image deblurring,''
\newblock in {\em Proceedings of the AAAI Conference on Artificial
  Intelligence}, 2023, vol.~37, pp. 1905--1913.

\bibitem{db3}
Syed~Waqas Zamir, Aditya Arora, Salman Khan, Munawar Hayat, Fahad~Shahbaz Khan,
  and Ming-Hsuan Yang,
\newblock ``Restormer: Efficient transformer for high-resolution image
  restoration,''
\newblock in {\em Proceedings of the IEEE/CVF conference on computer vision and
  pattern recognition}, 2022, pp. 5728--5739.

\bibitem{d5}
Abdelrahman Abdelhamed, Stephen Lin, and Michael~S Brown,
\newblock ``A high-quality denoising dataset for smartphone cameras,''
\newblock in {\em Proceedings of the IEEE conference on computer vision and
  pattern recognition}, 2018, pp. 1692--1700.

\bibitem{d7}
Liangyu Chen, Xiaojie Chu, Xiangyu Zhang, and Jian Sun,
\newblock ``Simple baselines for image restoration,''
\newblock in {\em European Conference on Computer Vision}. Springer, 2022, pp.
  17--33.

\bibitem{d8}
Alexander Kirillov, Eric Mintun, Nikhila Ravi, Hanzi Mao, Chloe Rolland, Laura
  Gustafson, Tete Xiao, Spencer Whitehead, Alexander~C Berg, Wan-Yen Lo,
  et~al.,
\newblock ``Segment anything,''
\newblock {\em arXiv preprint arXiv:2304.02643}, 2023.

\bibitem{hazing}
Zheyan Jin, Shiqi Chen, Yueting Chen, Zhihai Xu, and Huajun Feng,
\newblock ``Let segment anything help image dehaze,''
\newblock {\em arXiv preprint arXiv:2306.15870}, 2023.

\bibitem{realblur}
Jucheol~Won Jaesung~Rim, Haeyun~Lee and Sunghyun Cho,
\newblock ``Real-world blur dataset for learning and benchmarking deblurring
  algorithms,''
\newblock in {\em Proceedings of the European Conference on Computer Vision
  (ECCV)}, 2020.

\bibitem{relo}
Haoying Li, Ziran Zhang, Tingting Jiang, Peng Luo, Huajun Feng, and Zhihai Xu,
\newblock ``Real-world deep local motion deblurring,''
\newblock in {\em Proceedings of the AAAI Conference on Artificial
  Intelligence}, 2023, vol.~37, pp. 1314--1322.

\bibitem{unet}
Olaf Ronneberger, Philipp Fischer, and Thomas Brox,
\newblock ``U-net: Convolutional networks for biomedical image segmentation,''
\newblock in {\em Medical Image Computing and Computer-Assisted
  Intervention--MICCAI 2015: 18th International Conference, Munich, Germany,
  October 5-9, 2015, Proceedings, Part III 18}. Springer, 2015, pp. 234--241.

\bibitem{tlc}
Xiaojie Chu, Liangyu Chen, Chengpeng Chen, and Xin Lu,
\newblock ``Improving image restoration by revisiting global information
  aggregation,''
\newblock in {\em European Conference on Computer Vision}. Springer, 2022, pp.
  53--71.

\bibitem{psnr}
Alain Hore and Djemel Ziou,
\newblock ``Image quality metrics: Psnr vs. ssim,''
\newblock in {\em 2010 20th international conference on pattern recognition}.
  IEEE, 2010, pp. 2366--2369.

\end{thebibliography}

\end{document}